\documentclass[conference]{IEEEtran}
\IEEEoverridecommandlockouts

\usepackage{cite}
\usepackage{amsmath,amssymb,amsfonts}
\usepackage{algorithmic}
\usepackage{graphicx}
\usepackage{textcomp}
\usepackage{xcolor}
\usepackage{blindtext}
\def\BibTeX{{\rm B\kern-.05em{\sc i\kern-.025em b}\kern-.08em
    T\kern-.1667em\lower.7ex\hbox{E}\kern-.125emX}}

\makeatletter

\newcommand{\linebreakand}{%
  \end{@IEEEauthorhalign}
  \hfill\mbox{}\par
  \mbox{}\hfill\begin{@IEEEauthorhalign}
}

\def\ps@IEEEtitlepagestyle{%
  \def\@oddfoot{\mycopyrightnoticelow}%
  \def\@oddhead{\mycopyrightnoticetop}%
  \def\@evenhead{\@IEEEheaderstyle\thepage\hfil\leftmark\hbox{}}\relax
  \def\@evenfoot{}%
}

\def\mycopyrightnoticetop{%
  \begin{minipage}{\textwidth}
  \centering \scriptsize
  Bauer, J. C., Geng, P., Trattnig, S., Dokládal, P., \& Daub, R. (2025). Multi-Level Feature Fusion for Continual Learning in Visual Quality Inspection. In \textit{2025 IEEE 13th International Conference on Control, Mechatronics and Automation (ICCMA)} (pp. 587-592). IEEE. https://doi.org/10.1109/ICCMA67641.2025.11369688\\ Published version available at https://ieeexplore.ieee.org/document/11369688
  \end{minipage}
}

\def\mycopyrightnoticelow{%
  \begin{minipage}{\textwidth}
  \centering \scriptsize
  Copyright~\copyright~2025 IEEE. Personal use of this material is permitted. Permission from IEEE must be obtained for all other uses, in any current or future media, including reprinting/republishing this material for advertising or promotional purposes, creating new collective works, for resale or redistribution to servers or lists, or reuse of any copyrighted component of this work in other works.
  \end{minipage}
}

\makeatother

\makeatother

\begin{document}

\title{Multi-Level Feature Fusion for Continual Learning in Visual Quality Inspection\\
\thanks{The authors acknowledge funding by the German-French Academy for the Industry of the Future for the project "LAsSy - Learning Assistance Systems for Efficient Disassembly".
}
}

\author{
\IEEEauthorblockN{1\textsuperscript{st} Johannes C. Bauer}
\IEEEauthorblockA{\textit{Technical University of Munich,} \\
\textit{Institute for Machine Tools and} \\ \textit{Industrial Management (iwb)}\\
Garching, Germany \\
johannes.bauer@iwb.tum.de}
\and
\IEEEauthorblockN{2\textsuperscript{nd} Paul Geng}
\IEEEauthorblockA{\textit{Technical University of Munich,} \\
\textit{Institute for Machine Tools and} \\ \textit{Industrial Management (iwb)}\\
Garching, Germany \\
paul.geng@iwb.tum.de}
\and
\IEEEauthorblockN{3\textsuperscript{rd} Stephan Trattnig}
\IEEEauthorblockA{\textit{Technical University of Munich,} \\
\textit{Institute for Machine Tools and} \\ \textit{Industrial Management (iwb)}\\
Garching, Germany \\
stephan.trattnig@iwb.tum.de}
\linebreakand
\IEEEauthorblockN{4\textsuperscript{th} Petr Dokládal}
\IEEEauthorblockA{\textit{MINES Paris, PSL University,} \\
\textit{Centre for Mathematical Morphology (CMM)}\\
Fontainebleau, France \\
petr.dokladal@minesparis.psl.eu}
\and
\IEEEauthorblockN{5\textsuperscript{th} Rüdiger Daub}
\IEEEauthorblockA{\textit{Fraunhofer Institute for Casting,} \\
\textit{Composite and Processing Technology IGCV}\\
Augsburg, Germany \\
ruediger.daub@igcv.fraunhofer.de}
}

\maketitle

\begin{abstract}
Deep neural networks show great potential for automating various visual quality inspection tasks in manufacturing. However, their applicability is limited in more volatile scenarios, such as remanufacturing, where the inspected products and defect patterns often change. In such settings, deployed models require frequent adaptation to novel conditions, effectively posing a continual learning problem. To enable quick adaptation, the necessary training processes must be computationally efficient while still avoiding effects like catastrophic forgetting. 
This work presents a multi-level feature fusion (MLFF) approach that aims to improve both aspects simultaneously by utilizing representations from different depths of a pretrained network. We show that our approach is able to match the performance of end-to-end training for different quality inspection problems while using significantly less trainable parameters. Furthermore, it reduces catastrophic forgetting and improves generalization robustness to new product types or defects. 
\end{abstract}


\section{Introduction}
Deep neural network (DNN)-based quality inspection in manufacturing has been researched for many years~\cite{Wang.2018}. However, its application in scenarios with a high variety of product variants and small lot sizes still poses challenges. One example is remanufacturing, which refers to the process of restoring a used product to like-new condition~\cite{Lund.1983}. 
Since this preserves much of the originally created value, e.g., from raw material production, it offers economic and ecological benefits~\cite{Lund.1983}.
During the remanufacturing process, a product is disassembled into its components, which are cleaned and visually inspected for defects and signs of wear. This inspection determines whether a component can be directly reused, or must be refurbished or replaced before the product can be reassembled~\cite{Matsumoto.2016}. Thus, the visual inspection step has a high influence on both product quality and the economic viability of the process~\cite{Errington.2013}. Due to the many uncertainties, e.g., regarding time, quantity, and type of the returned products as well as the many different components that must be handled, this inspection is usually still done manually today~\cite{Errington.2013}.
Some scientific studies have already confirmed the potential of DNN for automating the inspection process by classifying or localizing defects~\cite{Nwankpa.2021},\cite{Saiz.2021},\cite{ Mohandas.2025}. However, these works focus on individual components. Due to the high component variety and uncertainties mentioned above, such DNN-based inspection systems will require continuous adaptation in practical applications.

To be able to perform those model adaptations frequently and/or on less powerful edge devices they should be computationally efficient without causing the model to forget much of its previous 'knowledge', usually referred to as catastrophic forgetting~\cite{Wang.2024b}. Continual learning methods aim to bridge exactly this trade-off~\cite{Wang.2024b}. An effective option are so-called rehearsal methods, where a small subset of previous samples is incorporated during adaptation with new data~\cite{Hsu.30.10.2018}. Many methods utilize latent representations of the models for this selection process~\cite{Harun.25.08.2023},\cite{Rebuffi.2017},\cite{Chaudhry.2019},\cite{Shim.31.08.2020}. However, since the feature extractor is usually trained as well, these representations must be recomputed after each adaptation. 

Therefore, it would be beneficial to freeze the feature extractor to be able to reuse previously computed latent representations. Although several studies demonstrate that catastrophic forgetting can be further reduced by preserving features or parts of a pretrained feature extractor~\cite{Zhang.31.12.2019},\cite{Shon.17.08.2022},\cite{Boschini.01.06.2022},\cite{Ostapenko.2022},\cite{Panos.2023},\cite{Requeima.2019},\cite{Cui.2025},\cite{Du.2024}, these methods either increase computational requirements or still modify the feature extractor and therefore the representations. Furthermore, working with features from a frozen feature extractor alone drastically reduces performance for many quality inspection tasks, as we also show in section~\ref{sec:matchingperformance}.

Therefore, in this work, a method that employs a pretrained feature extractor DNN, utilizing representations from different abstraction levels (depths of the network), is presented. Our experiments demonstrate its benefits regarding three main aspects. First, the approach can match the performance of end-to-end training on multiple quality inspection datasets while significantly reducing the number of trained parameters and, therefore, compute and memory requirements. Second, it improves continual learning performance by omitting the need for recomputing latent representations and by reducing catastrophic forgetting. Third, the results suggest that this method can also enhance a model's generalization robustness when classifying images from new product types or defects.

The remainder of this paper is structured as follows. We outline the fundamentals and related works in section~\ref{sec:fundamentals}. Then, the proposed multi-level feature fusion (MLFF) approach is presented in section~\ref{sec:method}. Subsequently, we report on the experiments in section~\ref{sec:experiments}, followed by the conclusion in section~\ref{sec:conclusion}.

\section{Fundamentals and related works}\label{sec:fundamentals}
This section first introduces fundamental principles and methods of continual learning. Afterward, related works on preserving pretrained features are discussed. 

\subsection{Fundamentals of continual learning} 
Continual learning refers to a system's ability to incrementally acquire and integrate new knowledge throughout its lifecycle, to be able to adapt to real-world dynamics ~\cite{Wang.2024b}. A central challenge is catastrophic forgetting, which refers to the phenomenon that learning a new task worsens a model's ability to perform previous tasks~\cite{Wang.2024b}. 
The goal of continuous learning methods is to achieve high performance on new tasks and maintain performance on old tasks, while using as few historical examples as possible.~\cite{Wang.2024b}. 
Depending on the continual learning scenario, new tasks may differ from old ones due to the introduction of new classes and/or altered input data (e.g., due to a distribution shift) and may come with distinct task labels or not~\cite{Wang.2024b}. 
 
Broadly, continual learning methods can be categorized into architectural-, regularization-, and rehearsal-based methods~\cite{Parisi.2019}. While architectural methods adapt a network's topology when learning new tasks to form task-specific regions~\cite{Parisi.2017},\cite{Rusu.2022}
, regularization-based methods try to avoid changes to model parameters that are important for previous tasks~\cite{Aljundi.2018},\cite{Pomponi.2020}. 
Rehearsal-based methods, in turn, save selected samples in a historic data buffer and add them to the new training data when learning a novel~task \cite{Harun.25.08.2023},\cite{Rebuffi.2017},\cite{Chaudhry.2019},\cite{Shim.31.08.2020},\cite{Lopez-Paz.2017g},\cite{Aljundi.2019j},\cite{Prabhu.2020},\cite{Yoon.02.06.2021}. 
In this work, we focus on rehearsal-based methods due to their conceptual simplicity and higher robustness to hyperparameter values. Moreover, simple rehearsal-based methods were found to outperform many more sophisticated approaches~\cite{Hsu.30.10.2018}, and for most practical applications, storage space is not the limiting factor~\cite{Prabhu.2023}. 

Rehearsal methods can be further grouped depending on the stored and reused content. While in so-called experience replay the plain inputs like images are used, latent rehearsal approaches store and reuse intermediate representations from a model~\cite{Wang.2024b, Harun.25.08.2023}.

A key question for rehearsal-based continual learning is which samples to select from the buffer when training.  
A  simple option is to randomly select samples in a class-balanced way \cite{Prabhu.2020} or follow a first-in, first-out approach~\cite{Lopez-Paz.2017g}. 
Other strategies use latent representations of the data, usually extracted before the last classification layer of the model. This includes the work of \cite{Rebuffi.2017}, where a so-called herding strategy is used to maintain a representative set of samples. 
Authors in~\cite{Chaudhry.2019} suggest prioritizing samples near the class mean. 
In contrast, Adversarial Shapley Value Experience Replay (ASER) selects samples that are both representative of their own class as well as close to the decision boundary with other classes in representation space using shapely values~\cite{Shim.31.08.2020}. The method Gradually Select Less Prototypical (GRASP) follows a similar approach by populating the buffer with samples near the respective class means first, but gradually adding more and more samples that are further away from the class means. Therefore, the selection probability of a sample is based on its cosine distance to the mean feature vector of its class~\cite{Harun.25.08.2023}. 
Other methods focus on the model and the loss. For example, 
by selecting the samples that have the highest loss value after initially training the model on the new data~\cite{Aljundi.2019j} or inspecting the gradients to select samples~\cite{Yoon.02.06.2021}. However, these are computationally costly and may even require a whole additional training run -- compute resources that could otherwise be spent on more training steps instead~\cite{Harun.25.08.2023}.   

\subsection{Preserving features for continual learning}
The positive impact of preserving pretrained features in a DNN on continual learning performance was confirmed in several works~\cite{Zhang.31.12.2019},\cite{Shon.17.08.2022},\cite{Boschini.01.06.2022},\cite{Ostapenko.2022},\cite{Panos.2023},\cite{Requeima.2019},\cite{Cui.2025},\cite{Du.2024}. These utilize different approaches for feature preservation and are discussed in the following.

Authors in~\cite{Zhang.31.12.2019} present a method called side-tuning, which uses a pretrained network with fixed parameters, trains another network in parallel, and adds their outputs.  
In~\cite{Shon.17.08.2022}, a linearization of a pretrained network is used instead. The features of the original pretrained model are added to those of the linearized model and processed by a subsequent classifier. Only the classifier and the linearized model are trained. 
Boschni et al.~\cite{Boschini.01.06.2022} also use a pretrained sibling network with fixed parameters in parallel to the application-specific DNN. During training, a regularizing loss term is used that minimizes the distance between intermediate features of both models.    

In contrast to the above-mentioned approaches, Ostapenko et al.~\cite{Ostapenko.2022} directly use the last latent representation extracted by a pretrained DNN for their experiments. A multi-layer perceptron (MLP) classifier is trained on top of these representations during downstream continual learning. In their experiments, models pretrained on broader data showed improved performance. 
Other works, like \cite{Panos.2023} or \cite{Requeima.2019} utilize so-called Feature-wise Linear Modulation (FiLM) layers~\cite{Perez.2018} to reduce catastrophic forgetting. These are inserted inside a DNN's structure to scale and shift the activations of the convolutional layers. During continual learning, the existing network's parameters are frozen and only the FiLM parameters are trained.
Cui et al.~\cite{Cui.2025} follow a similar approach and insert application-specific adapter layers into a pretrained vision transformer (ViT), which are trained to adapt the model to new tasks. To reduce catastrophic forgetting, a mixture of task-specific adapters is used. 
Authors in~\cite{Du.2024} use a frozen ViT as a feature extractor to generate class-prototype vectors for a nearest mean classifier. This then predicts a sample's class based on the distance of its representation vector to these class prototypes. 

Since methods like \cite{Panos.2023},\cite{Requeima.2019}, or \cite{Du.2024} adapt models to new tasks by inserting trainable parameters into a network's architecture, they also require recomputing latent representations for old samples after adaptation to use them for the selection of historic samples. Although no gradient backpropagation is necessary, this may take up considerable time. 
This could be avoided by storing and reusing previously computed latent representations. 
However, simply reusing the last representation of a pretrained model, as done in \cite{Ostapenko.2022}, does not work well for quality inspection use cases (this is also shown in section~\ref{sec:matchingperformance}). 
Using multiple models in parallel \cite{Zhang.31.12.2019},\cite{Shon.17.08.2022},\cite{Boschini.01.06.2022} is also not feasible in practice due to the increased computational effort.  

\section{Method}\label{sec:method}
\begin{figure}[t]
    \includegraphics[width=\linewidth]{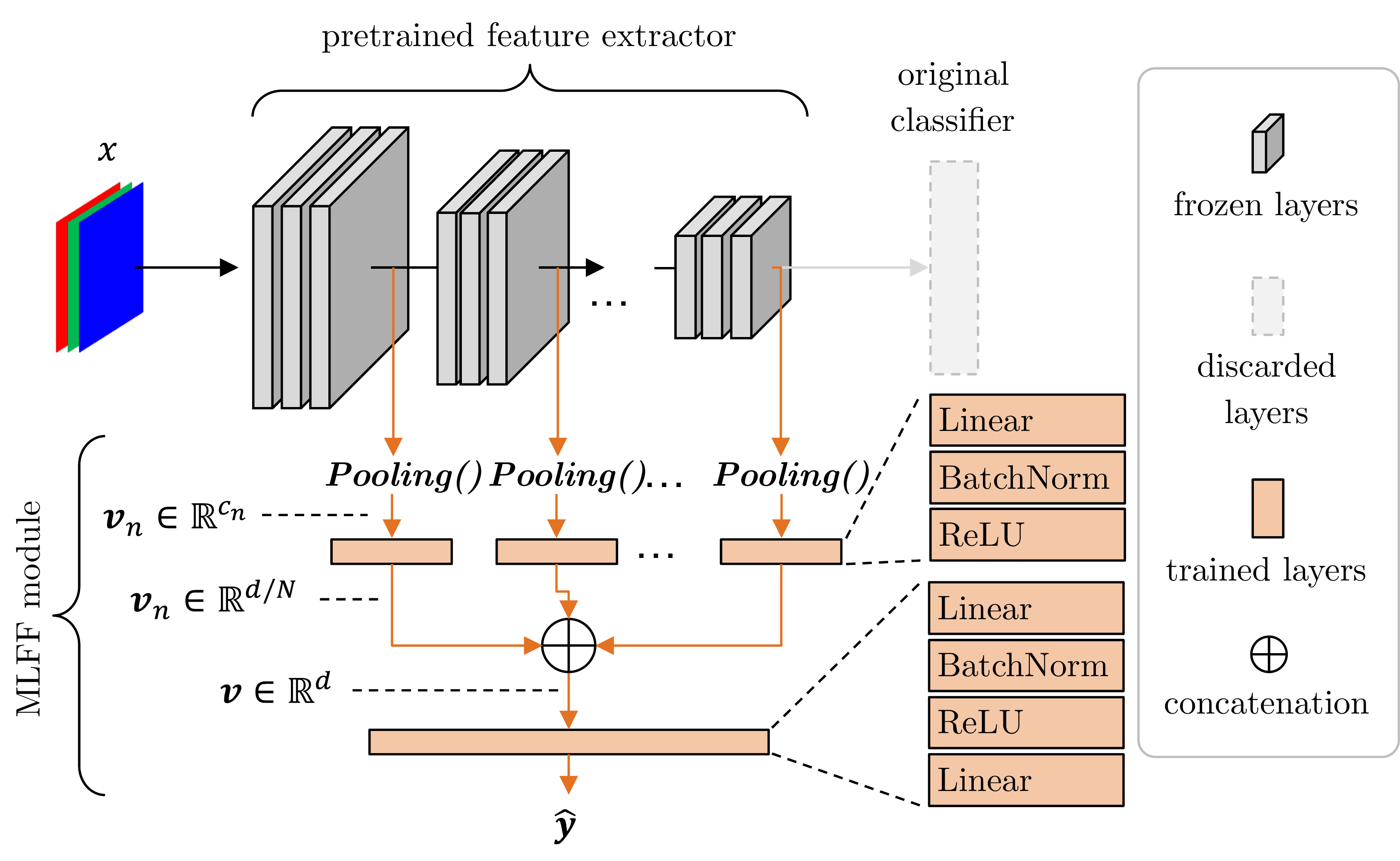}
    \caption{Overview of the proposed approach. As an input image $x$ is processed by the pretrained feature extractor, intermediate representations are extracted at different stages and processed by the MLFF module.}
    \label{fig:approach}
\end{figure}

The proposed approach addresses the shortcomings outlined above and aims to reduce computational effort and catastrophic forgetting by utilizing pretrained network layers, while still achieving competitive classification performance on quality inspection problems. To achieve this, we extract intermediate features from a pretrained backbone DNN at different abstraction levels and feed them to a lightweight classifier. During training, only the classifier's weights are optimized. 
Therefore, the representations extracted by the pretrained backbone network can be used to select historic samples in rehearsal-based continual learning without the need to recompute them after adapting the model to a new task. An overview of the approach is presented in Fig.~\ref{fig:approach} and explained below. 

The approach is conceptually simple and works with most common model architectures, both convolutional and transformer-based.  
The input image $\boldsymbol{x}$ is first processed by the \textit{pretrained feature} extractor, which is used to extract $N$ representation vectors $\boldsymbol{v}_1, ..., \boldsymbol{v}_n$ from different abstraction levels. Thereby, the extracted representations undergo separate \textit{pooling} operations. 
For spatial feature maps, the intermediate representations are average-pooled so their dimensionality corresponds to the number of channels $c_n$ at this stage. 
For ViT models, we concatenate the class token with the mean of the remaining tokens. 
These representations are used to select historic samples before they are further processed by the MLFF module (cf. section~\ref{sec:classessment}). 
Each representation vector is transformed by a separate linear layer, followed by batch normalization and ReLU (Rectified Linear Unit) activations. 
Thereby, their dimension is reduced to $d/N$, where $d$ is set to the dimensionality of the original model's embedding size before its classifier. Afterward, their outputs are concatenated to a single vector $\boldsymbol{v} \in \mathbb{R}^d$, which is processed by a two-layer MLP with hidden dimension $d$ that outputs the prediction~$\boldsymbol{\hat{y}}$. 

\section{Experiments}\label{sec:experiments}
In the following, the proposed method is experimentally evaluated. We first show that it is possible to match the classification performance of fully trained models using the MLFF approach in section~\ref{sec:matchingperformance}. Subsequently, its impact on continual learning performance and catastrophic forgetting is assessed in section~\ref{sec:classessment}. Our code is made available online\footnote{https://github.com/jhnns-br/mlff\_continual\_learning}. 

\subsection{Matching classification performance}\label{sec:matchingperformance}

\textit{\textbf{Implementation:}}
We use three publicly available image classification datasets resembling different quality inspection tasks. The Re-GBC dataset \cite{Bauer.2025b} contains over 15,000 images of different types of automotive transmission components in good and defective condition. The NEU-DET dataset \cite{Song.2013} contains 1,800 images of hot-rolled steel showing different defects, and the SDNET dataset \cite{Dorafshan.2018} focuses on crack detection and contains almost 60,000 images. 

The proposed MLFF approach is evaluated using different backbone models. These are ResNet50~\cite{He.2016}, MobileNetv3-S~\cite{Howard.06.05.2019}, DINOv2-B with registers~\cite{Darcet.2024}, and SwinV2-B~\cite{Liu.2022}. The number of intermediate representations $N$ is set to $4$ for all models, starting after the first block of layers and ending after the last block (before the original classifier).
All models are pretrained on ImageNet1k in a supervised fashion, except DINO, for which the weights from its self-supervised pretraining process are used~\cite{Caron.29.04.2021}. We use the Adam optimizer with a cosine annealing learning rate scheduler, tuning the learning rate and number of epochs for all models. The classification performance is measured using the F1-Score to be more robust to potential class imbalance. 
In the following, we compare the classification performance obtained by the MLFF approach, fully trained models, and finetuning of a linear or MLP classifier with frozen feature extractors.   

\begin{figure}[t]
    \includegraphics[width=\linewidth]{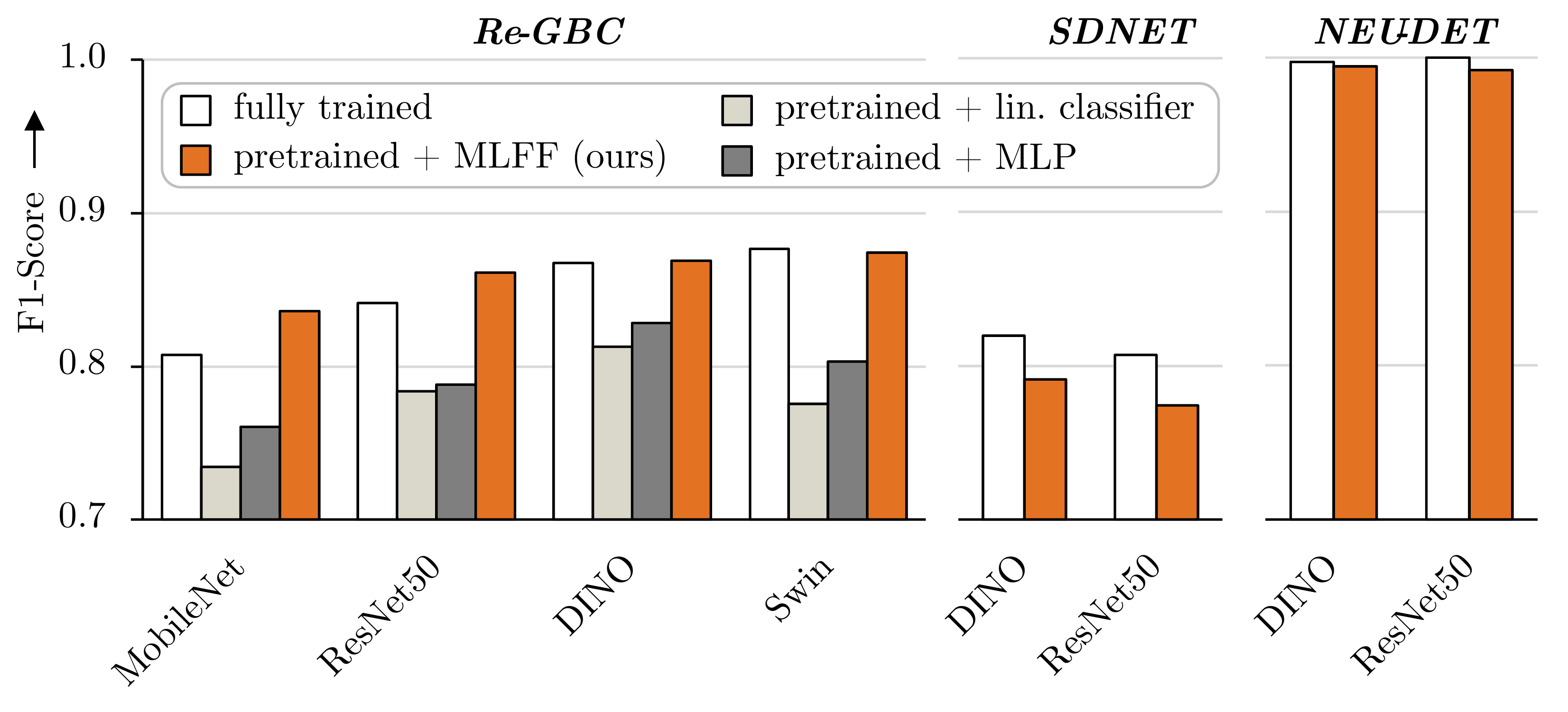}
    \caption{F1-Scores obtained on the test data of Re-GBC, SDNET, and NEU-DET using the MLFF approach, fully trained models, and finetuning of the linear and MLP classifiers.}
    \label{fig:matchingperformance}
\end{figure}

\textit{\textbf{Results:}}
The obtained classification performance on the respective datasets is visualized in Fig.~\ref{fig:matchingperformance}. The MLFF models (orange bars) are able to reach the same performance or even outperform the fully trained models (white bars) on the Re-GBC dataset. They also almost match performance for the SDNET dataset and NEU-DET, which suggests that the results are transferable over a wider range of quality inspection problems. 

Only training a linear or MLP classifier on top of a frozen feature extractor, which would correspond to the discarded classifier in Fig.~\ref{fig:approach}, (gray bars) is not sufficient to match the performance of the fully trained and MLFF models. A possible explanation is that common pretraining tasks favor more abstract feature representations, while in quality inspection, lower-level features like edges or rough surfaces are of more relevance. The latter may be obtained from the embeddings extracted at earlier stages in the network. 
This might also explain the slightly worse performance of the MLFF approach on the SDNET dataset. Some of the depicted cracks are very thin, and detecting them on downsampled feature maps may be difficult. This could be solved by extracting representations from even earlier stages.

At the same time, as shown in Table~\ref{tab:parameters}, the MLFF versions of ResNet, DINO, and Swin require training only 2 to 10~\% of the parameters of the fully trained versions. This significantly reduces computational effort and memory requirements.
The effect does not apply to the MobileNet architecture, which is already very lightweight. However, those models also perform worse than the larger transformer models DINO or Swin. 

\begin{table}[t]
    \caption{Number of parameters of the evaluated models.  
    }\label{tab:parameters}
    \centering
    \begin{tabular}{l c c c c }
    \hline
    & \multicolumn{4}{c}{\textbf{No. of parameters in million (overall/trained)}} \\
    \textbf{Configuration} & \textbf{\textit{ResNet50}} & \textbf{\textit{MobileNet}} & \textbf{\textit{DINO}} & \textbf{\textit{Swin}} \\
    \hline
    fully trained & 23.5/23.5 & 2.2/2.2 & 86.6/86.6 & 86.9/86.9 \\
    pretr. + lin. cl. & 23.5/0.004 & 2.2/0.003 & 86.6/0.003 & 86.9/0.002 \\
    pretr. + MLP & 27.7/4.2 & 3.9/1.6 & 89.0/2.4 & 88.0/1.1 \\
    pretr. + MLFF & 25.5/2 & 4.3/2.1 & 91.3/4.7 & 88.5/1.5 \\
    \hline
    \end{tabular}
\end{table}

\subsection{Continual learning performance}\label{sec:classessment}

\textit{\textbf{Implementation:}}
For the continual learning experiments, we utilize the Re-GBC dataset and partition it into five groups/tasks based on the provided part IDs. From each group, 1,500 randomly selected images are used for training, while the others are used for testing. The MLFF module is initially trained on the first task's training data and subsequently adapted four times using rehearsal-based continual learning. Thereby, three training epochs are performed for the new task's samples, during which we randomly mix in the historic samples (so each historic sample is only seen once). This was found to be sufficient for classification performance on the current task to converge. Basic data augmentation, like random gaussian noise and horizontal flips, is performed during all training processes.  

We report the \textit{average F1-Score} across all tasks after the final adaptation round (AF1) \cite{Lopez-Paz.2017g}. Additionally, we assess the generalization capability by computing the average F1-Scores achieved on the test data of all \textit{future tasks} after the respective adaptation rounds (FF1), excluding the final round, naturally.

We first compare fully trained models with the MLFF approach using random class-balanced sample selection~\cite{Prabhu.2020} since it is known to be a strong baseline \cite{Prabhu.2023}. Subsequently, additional rehearsal strategies are evaluated in combination with the MLFF architecture, incorporating the approaches below. These use the concatenated representation vectors after the pooling step to select historic samples. 

\begin{itemize}
    \item \textbf{farthest point sampling (FPS)} in representation space for each class
    \item \textbf{mean} as introduced in~\cite{Chaudhry.2019}.
    \item \textbf{GRASP} as introduced in~\cite{Harun.25.08.2023}.
    \item \textbf{ASER} as introduced in~\cite{Shim.31.08.2020}. 
\end{itemize}

\begin{figure*}[t]
    \includegraphics[width=\linewidth]{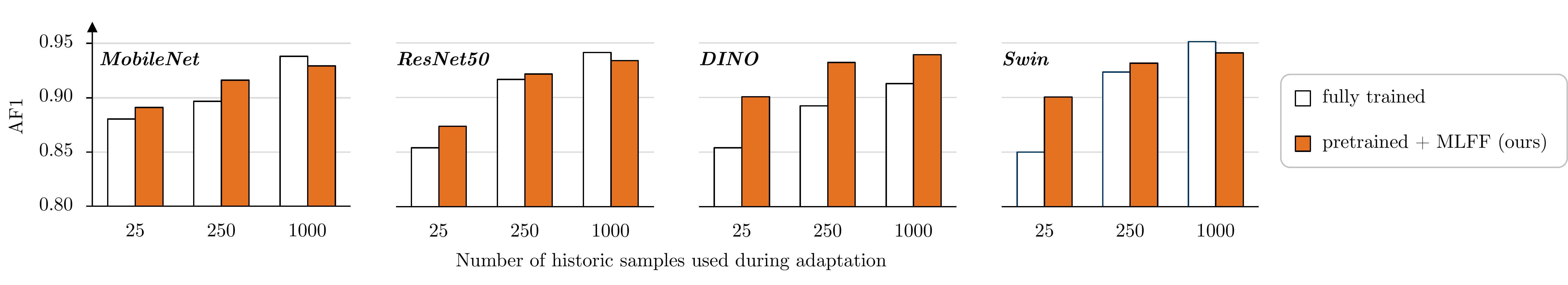}
    \caption{AF1 after continual learning on the Re-GBC dataset for different model architectures and numbers of historic samples.
    }
    \label{fig:clarchitecture}
\end{figure*}

\textit{\textbf{Results:}}
Fig.~\ref{fig:clarchitecture} visualizes the AF1 obtained on all five tasks of the evaluation scenario after four adaptation rounds. Especially if fewer historic samples are used, the MLFF approach significantly improves performance compared to the fully trained models. 
The positive effect is particularly pronounced for DINO, where the MLFF models show less catastrophic forgetting than the fully trained ones for all three scenarios, which might be caused by its ViT architecture. For the other three architectures, the fully trained versions still perform slightly better for 1,000 historic samples, although the differences are rather small, ranging up to 1 percentage point. Moreover, the MLFF approach facilitates using more historic samples by reducing the computational effort. In particularly for DINO and Swin, only a small fraction of parameters, compared to the original models, is trained (cf. section~\ref{sec:matchingperformance}).

Regarding the generalization performance, quantified by the FF1, we can also see a positive impact of the MLFF approach. As can be seen in Table~\ref{tab:generalization}, the generalization performance is improved for all model architectures on average. Especially for DINO and MobileNet, the improvement is quite significant, ranging from 6 to 11 percentage points. 

\begin{figure}[t]
    \includegraphics[width=\linewidth]{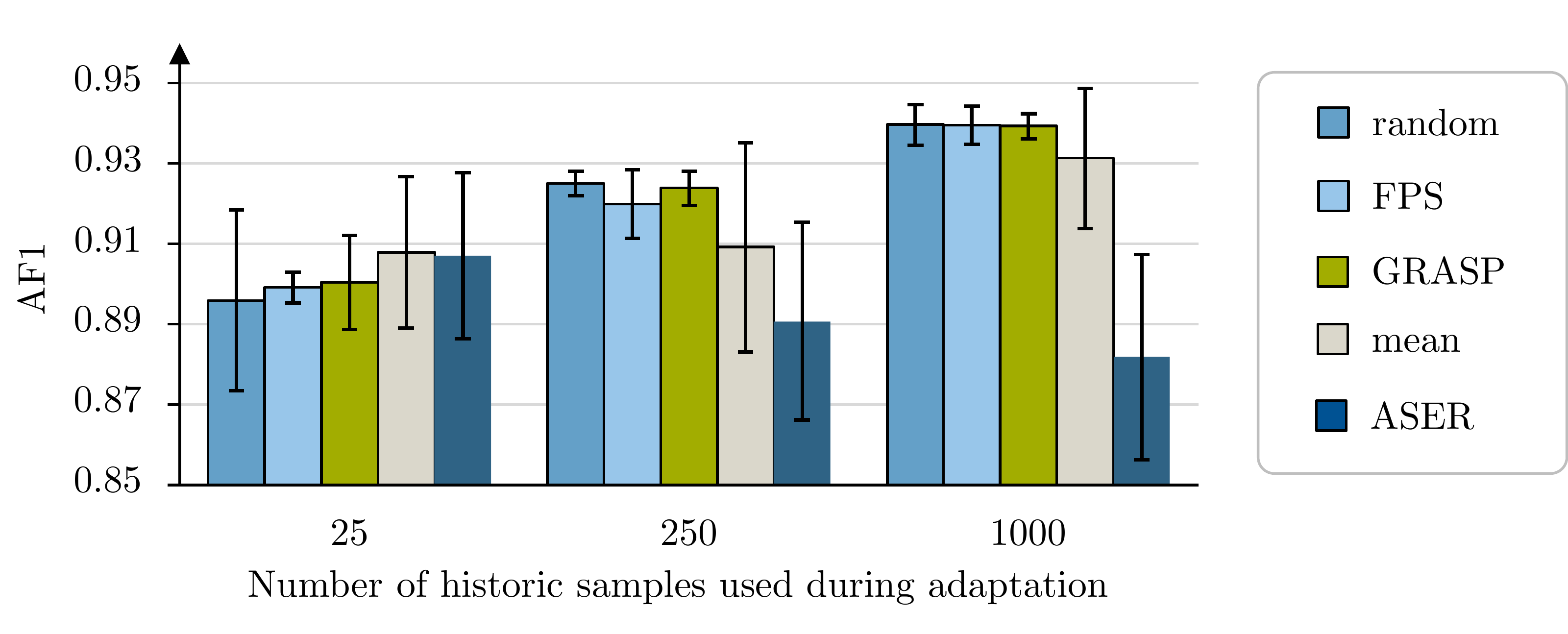}
    \caption{AF1 (+/- one standard deviation) obtained with a ResNet50 and the MLFF approach on the Re-GBC dataset for different rehearsal methods.}
    \label{fig:clmethoda}
\end{figure}

\begin{table}[t]
    \caption{FF1 scores when using fully trained (FT) and MLFF models. Higher values per model architecture and buffer size are marked \textbf{bold}. 
    }\label{tab:generalization}
    \centering
    \begin{tabular}{l|c c |c c| c c }
    \hline
         & \multicolumn{2}{c|}{\textbf{25}} & \multicolumn{2}{c|}{\textbf{250}} & \multicolumn{2}{c}{\textbf{1000}} \\
        \textbf{Architecture} & \textbf{\textit{FT}} & \textbf{\textit{MLFF}} & \textbf{\textit{FT}} & \textbf{\textit{MLFF}} & \textbf{\textit{FT}} & \textbf{\textit{MLFF}} \\ \hline
        MobileNet & 0.656 & \textbf{0.770} & 0.665 & \textbf{0.766} & 0.675 & \textbf{0.753} \\ 
        ResNet50  & 0.734 & \textbf{0.764} & 0.737 & \textbf{0.769} & \textbf{0.758} & 0.745 \\ 
        DINO      & 0.706 & \textbf{0.777} & 0.718 & \textbf{0.783} & 0.667 & \textbf{0.778} \\ 
        Swin      & \textbf{0.761} & 0.754 & 0.752 & \textbf{0.760} & 0.752 & \textbf{0.760} \\ \hline
    \end{tabular}
\end{table}

\begin{figure}[t]
    \includegraphics[width=\linewidth]{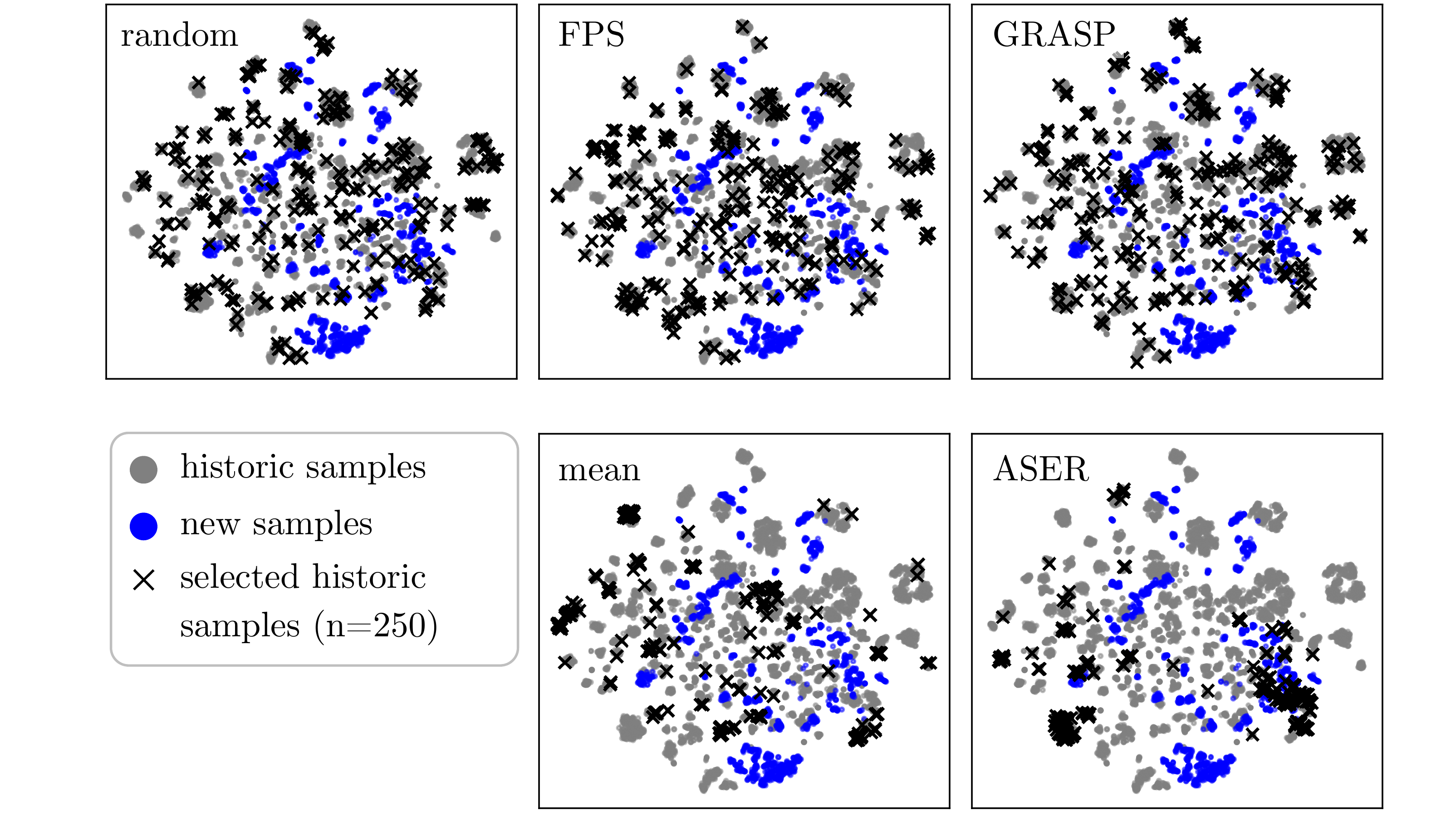}
    \caption{t-SNE embeddings of the new, historic, and selected historic samples during the last adaptation round for different rehearsal methods using a ResNet50 and the MLFF approach (best viewed in color).}
    \label{fig:clmethodb}
\end{figure}

Fig.~\ref{fig:clmethoda} shows results obtained by different rehearsal methods, when combined with a ResNet50 and the MLFF approach. Due to the small differences between the methods, the mean and standard deviation of three runs with different task sequences are reported. While approaches like Mean and ASER work better with smaller buffer sizes, approaches like random class-balanced, FPS, or GRASP outperform them for larger numbers of historic samples. This is in line with literature like \cite{Harun.25.08.2023} or \cite{Chaudhry.2019}, which point out that the selection of more class-typical samples by the mean-based approaches is beneficial in settings with few training samples. However, if more samples are used, approaches that yield a broader distribution of historic samples in representation space are favorable. This is underlined by Fig.~\ref{fig:clmethodb}, which shows t-SNE (t-distributed Stochastic Neighbor Embedding)~\cite{Maaten.2008} reduced representations of selected, historic, and novel samples. 
Overall, the differences between the improvements obtained by individual rehearsal methods over random class-balanced selection are rather small, compared to the improvement obtained by the MLFF approach. 

\section{Conclusion}\label{sec:conclusion}
The approach presented in this work utilizes latent representations from different depths of a pretrained DNN to enhance robustness and reduce catastrophic forgetting in continual learning scenarios in quality inspection. We show that the approach can achieve competitive performance on multiple quality inspection problems without requiring application-specific training of the feature extractor. This significantly reduces the number of trained parameters (down to 2~\% for some models) and necessary computational effort. 
Further experiments confirm its effectiveness in mitigating catastrophic forgetting and improving generalization performance in rehearsal-based continual learning, i.e., when only a few historic samples are used. 

There are several directions for further research. First, the approach could be applied to additional model architectures to further assess its generalizability. Second, the integration of other mechanisms like attention could be explored, alongside variations in the number and position of layers from which features are extracted. Lastly, the incorporation of continual learning techniques like Dark Experience Replay \cite{Buzzega.15.04.2020} could be investigated to further reduce catastrophic forgetting.

\section*{Author Contributions} 
JCB: Conceptualization, Methodology, Software, Investigation, Visualization, Writing - Original Draft, Writing - Review \& Editing, Project administration. PG: Methodology, Writing - Review \& Editing. ST: Methodology, Writing - Review \& Editing. PD: Methodology, Supervision, Writing - Review \& Editing. RD: Resources, Supervision, Funding Acquisition, Writing - Review \& Editing.

\bibliographystyle{ieeetr}
\bibliography{ieee-bibliography}{}
\end{document}